\documentclass{article}
\usepackage[utf8]{inputenc}
\usepackage{authblk}

\title{Active Fairness Instead of Unawareness}

\author{Boris Ruf}
\author{Marcin Detyniecki\thanks{\{boris.ruf,marcin.detyniecki\}@axa.com}}
\affil{AXA Research, Paris, France}

\date{February 2020}

\begin{document}

\maketitle

\begin{abstract}
The possible risk that AI systems could promote discrimination by reproducing and enforcing unwanted bias in data has been broadly discussed in research and society. Many current legal standards demand to remove sensitive attributes from data in order to achieve ``fairness through unawareness". We argue that this approach is obsolete in the era of big data where large datasets with highly correlated attributes are common. In the contrary, we propose the active use of sensitive attributes with the purpose of observing and controlling any kind of discrimination, and thus leading to fair results.
\end{abstract}

\section{Current regulations to avoid discrimination}

Systematic, unequal treatment of individuals based on their membership of a sensitive group is considered discrimination. There is broad consensus in our society that it is unfair to make a distinction on the ground of a personal characteristic which is usually not a matter of choice. Therefore, most legal frameworks prohibit such actions. When it comes to non-discrimination in the EU, for example, the \emph{Convention for the Protection of Human Rights and Fundamental Freedoms} defines the ``Prohibition of discrimination" in Article 14~\cite{Europe:1950uq}. This principle is further contained in the \emph{Charter of Fundamental Rights of the European Union} which states in Article 21 that ``Any discrimination based on any ground such as sex, race, colour, ethnic or social origin, genetic features, language, religion or belief, political or any other opinion, membership of a national minority, property, birth, disability, age or sexual orientation shall be prohibited."~\cite{EUCharter}

In automatic decision-making, the traditional approach to fight discrimination is known as ``anti-classification" among legal scholars~\cite{Corbett2018}. This principle attempts to obtain ``fairness through unawareness" by simply excluding any sensitive attributes as features from the data. In EU law, this is enforced on the level of data protection: The General Data Protection Regulation (GDPR) regulates the collection and use of personal data, including sensitive personal data. For many use cases, it strictly prohibits the storing and processing of a list of attributes which were classified as protected, i.e. sensitive~\cite{GDPR}.

For non-AI systems, when using conventional, deterministic algorithms with a manageable amount of data, the current approach can provide a solution. However, it is important to point out that in the case of ill intention, anti-classification does not prevent discrimination per se, as the practice called ``redlining" has proven in the past: Sometimes, non-sensitive attributes may be strongly linked to sensitive attributes. Consequently, they can serve as substitutes or proxies. For example, the non-sensitive attribute zip code might be correlated with the sensitive attribute race when many people from the same ethnic background live in the same neighbourhood. Hence, already in the context of non-AI systems, seemingly unsuspicious attributes can be misused to produce discriminatory decisions with the purpose of explicitly excluding a specific sensitive subgroup~\cite{Gano2017}. Without actual knowledge of the sensitive attributes, such actions are hard to detect and to prevent.

\section{Why this is totally insufficient for AI systems}

When it comes to AI systems, the concept of removing sensitive attributes from data in order to prevent algorithms from being unfair has proven particularly insufficient: Such systems are usually backed by high-dimensional and strongly correlated datasets. This means that the decisions are based on hundreds or even thousands of attributes whose relevance is not obvious at first glance for the human eye. Further, some of those attributes usually contain strong links which again are difficult to spot for humans. Even after removing the sensitive attributes, such complex correlations in the data may continue to provide many unexpected links to protected information. In fact, heuristic methods exist to actively reconstruct missing sensitive attributes. For example, the Bayesian Improved Surname Geocoding (BISG) method attempts to predict the race given the surname and a geolocation~\cite{Elliott2008}. While the reliability of this method is generally disputed, it demonstrates that prohibiting to collect sensitive attributes does not prevent any possible misuse just by technical design. 

But even without any bad intent to discriminate, there is a danger of hidden indirect discrimination which is very difficult to detect in the results~\cite{Hajian2013,Corbett2018}. To illustrate the problem, we imagine an AI system which analyses CVs in order to propose starting salaries for newly hired staff. We further assume that women were discriminated in the past because their salaries were systematically lower compared to those of their male colleagues. Historical bias of this kind cannot be overcome by excluding the sensitive attribute ``gender" in the learning data since many links to non-sensitive attributes exist. For example, some hobbies or sports may be more popular among women or men. In languages with grammatical gender, the applicant's gender may be revealed through gender inflections of nouns, pronouns or adjectives. And yet more complex, in a country with compulsory military service exclusively for men, the entry age at university could provide a hint to the gender, too. Even when trying to additionally adjust for all of those identified correlations manually, it remains impossible to establish a sufficient degree of ''unawareness" which could guarantee discrimination-free decisions.

\section{Our recommendation: Active fairness}
We acknowledge the risk to privacy protection when storing sensitive attributes and we understand the motivation behind the current rules to address those concerns. However, based on the considerations in the previous sections, we conclude that the current practice of trying to ignore the existence of sensitive subgroups by omitting sensitive attributes bares greater risk than any privacy concerns related to the data collection. We therefore suggest to re-examine the status quo and propose the \textbf{active use of sensitive attributes in AI systems} to make sensitive subgroups visible and account for them with the purpose of verifiable fair results. Such a paradigm shift would allow for statistical measures to detect any type of discrimination and make it possible to mitigate unwanted bias in the data. Allowing to collect the sensitive attributes would be a helpful step towards verification mechanisms for AI stakeholders to test for imbalanced results, as well as for third parties such as regulators who could audit the data to ensure non-discrimination of underprivileged subgroups. To protect the sensitive attributes from misuse, new technical security mechanisms such as restricted access rules could be established.

In a nutshell, the principle of ``active fairness" as opposed to ``fairness through unawareness" would lay the foundations for tools which ensure that the standards of fairness and non-discrimination in AI systems are respected. Ultimately, access to such instruments would clear the way for increased trust in AI systems in society and could contribute to overcome a problem which has plagued humanity ever since: human bias. Fixing biases in algorithms remains a technical problem which is complex but still far easier to solve than correcting cognitive bias. If we succeed in developing automated systems which can help us taking fair, impartial decisions, the potential contribution for human progress and the protection and support of disadvantaged groups will be enormous.

\bibliographystyle{unsrt}
\bibliography{sample}
\end{document}